%% file: tnnls_2023.tex
\documentclass[lettersize,journal]{IEEEtran}
\usepackage{amsmath,amsfonts}
\usepackage{algorithmic}
\usepackage{algorithm}
\usepackage{array}
\usepackage[caption=false,font=normalsize,labelfont=sf,textfont=sf]{subfig}
\usepackage{textcomp}
\usepackage{stfloats}
\usepackage{url}
\usepackage{verbatim}
\usepackage{graphicx}
\usepackage{cite}

\usepackage{xcolor}
\usepackage{amsmath}
\usepackage{tabularx}
\usepackage{multirow}
\usepackage{booktabs}

\usepackage{amsbsy}
\newcommand\encoder{\#Encoder}
\begin{document}

% \title{A Sample Article Using IEEEtran.cls\\ for IEEE Journals and Transactions}

\title{HICL: Hashtag-Driven In-Context Learning for Social Media Natural Language Understanding}

% \author{IEEE Publication Technology,~\IEEEmembership{Staff,~IEEE,}
%         % <-this % stops a space
% \thanks{This paper was produced by the IEEE Publication Technology Group. They are in Piscataway, NJ.}% <-this % stops a space
% \thanks{Manuscript received April 19, 2021; revised August 16, 2021.}}

% \author{
% Hanzhuo Tan,~\IEEEmembership{Student Member,~IEEE}, Li Jing,~\IEEEmembership{Member,~IEEE}

\author{
Hanzhuo Tan, Chunpu Xu, Jing Li, Yuqun Zhang,~\IEEEmembership{Member,~IEEE}, Zeyang Fang, Zeyu Chen, Baohua Lai

\thanks{This work is supported by the NSFC Young Scientists Fund (No.62006203), a grant from the Research Grants Council of the Hong Kong Special Administrative Region, China (Project No. PolyU/25200821), the Innovation and Technology Fund (Project No. PRP/047/22FX), and CCF-Baidu Open Research Fund (No. 2021PP15002000). \emph{(Corresponding author: Jing Li.)}}
\thanks{Hanzhuo Tan is with the Department of Computing, Hong Kong Polytechnic University, Hong Kong, and with the Department of Computer Science and Engineering, Southern University of Science and Technology, Shenzhen, China (email: hanzhuo.tan@connect.polyu.hk).}
\thanks{Chunpu Xu and Jing Li are with the Department of Computing, Hong Kong Polytechnic University, Hong Kong (email: chun-pu.xu@connect.polyu.hk; jing-amelia.li@polyu.edu.hk).}
% \thanks{Jing Li is with the Department of Computing, the Hong Kong Polytechnic University, Hong Kong (email: jing-amelia.li@polyu.edu.hk).}
\thanks{Yuqun Zhang is with the Department of Computer Science and Engineering, Southern University of Science and Technology, Shenzhen (email: zhangyq@sustech.edu.cn).}
\thanks{Zeyang Fang, Zeyu Chen, and Baohua Lai are with Baidu Inc. (email: fangzeyang@baidu.com; chenzeyu01@baidu.com; laibaohua@baidu.com).}
% \thanks{Zeyu Chen is with Baidu Inc. (email: chenzeyu01@baidu.com).}
% \thanks{Baohua Lai is with Baidu Inc. (email: laibaohua@baidu.com).}
}

% The paper headers
% \markboth{Journal of \LaTeX\ Class Files,~Vol.~14, No.~8, August~2021}%
% {Shell \MakeLowercase{\textit{et al.}}: A Sample Article Using IEEEtran.cls for IEEE Journals}

% \markboth{IEEE Transactions on Neural Networks and Learning Systems}%
% {Shell \MakeLowercase{\textit{Tan et al.}}: HICL: Hashtag-Driven In-Context Learning for Social Media Natural Language Understanding}

% \IEEEpubid{0000--0000/00\$00.00~\copyright~2021 IEEE}
% Remember, if you use this you must call \IEEEpubidadjcol in the second
% column for its text to clear the IEEEpubid mark.

\maketitle

\begin{abstract}

Natural language understanding (NLU) is integral to various social media applications.
However, existing NLU models rely heavily on context for semantic learning, resulting in compromised performance when faced with short and noisy social media content. 
To address this issue, we leverage in-context learning (ICL), wherein language models learn to make inferences by conditioning on a handful of demonstrations to enrich the context and propose a novel hashtag-driven in-context learning (HICL) framework.
% Drawing motivation from the success of in-context learning (ICL), wherein language models learn to perform inferences by conditioning on a handful of demonstrations, we propose an approach based on hashtags for in-context social media language understanding. 
Concretely, we pre-train a model \encoder{}, which employs \#hashtags (user-annotated topic labels) to drive BERT-based pre-training through contrastive learning. 
% We aim to enhance the capacity of \encoder to incorporate topic-related semantic information, enabling it to retrieve topic-related posts later and improve contextual understanding for social media NLU (namely \textit{in-context social media NLU}). 
Our objective here is to enable \encoder{} to gain the ability to incorporate topic-related semantic information, which allows it to retrieve topic-related posts to enrich contexts and enhance social media NLU with noisy contexts.
To further integrate the retrieved context with the source text, we employ a gradient-based method to identify trigger terms useful in fusing information from both sources.
For empirical studies, we collected 45M tweets to set up an in-context NLU benchmark, and the experimental results on seven downstream tasks show that HICL substantially advances the previous state-of-the-art results.
Furthermore, we conducted extensive analyzes and found that: (1) combining source input with a top-retrieved post from \encoder{} is more effective than using semantically similar posts; 
(2) trigger words can largely benefit in merging context from the source and retrieved posts.

% Our results also indicate that trigger words are essential for information integration between source and retrieved posts. We present further quantitative analysis to interpret how hashtag-gathered posts advance generic NLU against data sparsity.

\end{abstract}

\begin{IEEEkeywords}
Nature language processing, social media, pre-trained language model, in-context learning.
\end{IEEEkeywords}

\input{1Introduction}

\input{2Related-work.tex}
\input{4Enrich.tex}
\input{5Experiment-setup.tex}

\input{6Experiment-results.tex}
\input{7Conclusions.tex}

\bibliography{custom,anthology}
\bibliographystyle{IEEEtran}

\end{document}

%% file: 1Introduction.tex
\section{Introduction}

\IEEEPARstart{S}{ocial} media provides rich resources of real-life, real-time content to understand our world and society. 
It motivates the demands and advances of various NLP applications on there, such as stance detection \cite{glandt-etal-2021-stance} and content recommendation \cite{zeng-etal-2020-dynamic}.
For these applications, natural language understanding (NLU) plays an essential role in featuring the text and representing its semantics, where pre-trained language models \cite{devlin-etal-2019-bert,roberta,T5} contribute cutting-edge advances and serve as the backbone to broadly benefit downstream applications \cite{barbieri-etal-2020-tweeteval}.

Pre-trained language models gain generic NLU capabilities by navigating large-scale text and exploring the context, such as word co-occurrence patterns.
Their performance thus heavily relies on \textit{rich and high-quality} context, whereas that on social media is prevalently short and noisy. 
It results in a severe problem of \textit{data sparsity}, meaning the context on social media exhibits an extremely sparse distribution of language features \cite{zeng-etal-2018-topic}. 
It would universally and negatively affect NLU pre-training and its downstream tasks \cite{knowledge-classification, knowlege-match}.
Viewing this challenge, BERTweet and Bernice are pre-trained  by randomly concatenating social media posts to lengthen context and alleviate sparsity \cite{nguyen-etal-2020-bertweet, delucia-etal-2022-bernice}.
It is, however, suboptimal as random  are unlikely to form coherent contexts.

Given these concerns, we envision that effective methods to automate \textit{context enriching} will allow less sparse features and promisingly advance the generic NLU on social media.
Our idea is inspired by recent advances in \textit{in-context learning} (ICL) %\cite{DBLP:journals/corr/abs-2207-10062}
\cite{min-etal-2022-metaicl,min2021noisy,min2022rethinking}, uplifting model performance via conditioning on a few example data from training samples. 
%
% However, the existing ICL research exhibits two-fold concerns.
% On the one hand, prior efforts are predominantly made on uni-directional models like GPT \cite{brown2020language} and have primarily overlooked bi-directional models, such as the BERT family, despite the unique advantages of the latter.
%
% However, the existing ICL researches are predominantly done on uni-directional models like GPT \cite{brown2020language} and have primarily overlooked bi-directional models, such as the BERT family, despite the unique advantages of the latter.
% Its bidirectional attention mechanisms can incorporate context from both directions when encoding a word or sentence, and hence
% demonstrate superior performance on NLU tasks \cite{devlin-etal-2019-bert}.
% This way, models can effectively capture linguistic phenomena, such as long-distance dependencies, pronoun resolution, and negation understanding.
% It also reflects how human readers process language since we understand words and sentences beyond solely relying on left-to-right contexts because it cannot fully capture the dependencies between the context words \cite{du-etal-2022-glm}.
%
However, the existing ICL researches are predominantly done on uni-directional models like GPT \cite{brown2020language} and have primarily overlooked bi-directional models, such as the BERT family, despite the unique advantages of the latter on NLU tasks \cite{devlin-etal-2019-bert}.

Motivated by the above points, our study aims to effectively retrieve external data and properly fine-tune bi-directional models to advance generic NLU on social media (henceforth \textbf{in-context social media NLU}). 
To that end, we first pre-train an embedding model to help any social media post in context enriching by retrieving another relevant post; then, we insert trigger terms to fuse the enriched context for language models to refer to in semantics learning under sparsity. 
This way, the framework can easily be plugged into various task-specific fine-tuning frameworks as external features and broadly benefits downstream social media tasks.

In existing approaches, ICL examples are usually constructed by retrieving the samples using metrics like semantic similarity \cite{liu-etal-2022-makes} and mutual information \cite{sorensen-etal-2022-information}. 
However, its effectiveness is concerning due to social media's short and informally-written text.
A related study about social media image-text understanding \cite{xu-and-li-2022-borrowing} showed retrieving context-enriching data was helpful; yet, image features contribute substantially more than text in retrieval training.
It sheds light on the non-trivial challenge of learning what to retrieve given data sparsity in a text-only context, which is much more common in social media posts than those with images. 
%limited and fragmented text context in a tweet. 
To address this issue, we pre-train the retrieval model via utilizing \textit{hashtags}, user-annotated topic labels starting with a ``\#'' and cross-referring to other topic-related posts \cite{zhang-etal-2021-howyoutagtweets}.
It associates posts about the same topic, learns semantics in a richer topic-coherent context, and gains topic relevance for retrieval.
Hashtags were adopted in many task-specific scenarios (e.g., image captioning \cite{lu-etal-2018-entity} and sentiment analysis \cite{gyanendro-singh-etal-2020-sentiment}). 
In contrast, we present a novel initiative to explore its effects in ICL for a broad range of NLU tasks.

\input{figure/intro-case.tex}

To better illustrate the potential of hashtags in context-enriching, Figure \ref{intro-case} shows a sample tweet $P$ conveying a sense of sarcasm through an emoticon ``\textit{zzzZZZ}'' (indicating overwork and opposing the previous sayings).
As can be seen, $P$'s short context and implicit writings may hinder NLU models from capturing the genuine underlying meanings.
We then retrieved a tweet $S$ using a popular semantic-based retrieval model SimCSE \cite{gao-etal-2021-simcse}, which exhibits similar semantics (heavy work), partially helps enrich context, yet ignores the sarcastic hint from ``\textit{zzzZZZ}''.
Meanwhile, a related hashtag ``\textit{\#zzz}'' might gather other topic-related  (like $T$ in Figure \ref{intro-case}) complaining about overwork, strengthen the NLU in ``\textit{zzzZZZ}'', and offer more direct assistance to infer sarcasm.

Here a straightforward approach is to feed the encoder with the concatenation of the source post with another topic-related post.
Nevertheless, this method may also distract the model, causing it to pay undue attention to non-essential details instead of focusing on the main message of the source post.
% We additionally observe that immediately combining the original tweet with a topic-related tweet may cause the model to be distracted,  
Therefore, we employ a gradient-based approach to identify trigger terms that facilitate the incorporation of the retrieved text's context.
To the best of our knowledge, \textit{Hashtag-Driven In-Context Learning (HICL) is the first framework leveraging hashtags in large-scale pre-training for social media NLU, which enables the pre-trained model to retrieve topic-related posts and enhances the ICL framework by incorporating automatically generated trigger terms for context enrichment.
%and advancing broad social media tasks. 
%for context-enriching with the help of automatically generated trigger terms and advancing broad social media tasks. 
}
%To the best of our knowledge, \textit{our hashtag-driven in-context learning (HICL) is the first model on social media NLU, where hashtags help in developing the first topic-aware \textbf{\encoder{}} through large-scale language pre-training.
%}
%and present the first topic-aware \encoder{} through large-scale language pre-training. 
%Experimental results demonstrate that the HICL framework effectively advances social media tasks. 
%}
% To the best of our knowledge, \textit{we are the first to leverage hashtags in large-scale language pre-training and enable it to retrieve topic-related tweets for context-enriching with the help of automatically generated trigger terms and advancing broad social media tasks. }

Concretely, HCL works in a pre-training and fine-tuning paradigm. 
In pre-training, we develop \textit{\encoder{}}, a hashtag-driven pre-trained model based on RoBERTa \cite{roberta}.
It is pre-trained on 179 million hashtagged tweets via contrastive learning to pull the tweets with the same hashtags closer together in embedding space and push apart those with different hashtags.
%\textit{HICL}, a Hashtag-Driven In-Context Learning framework for social media NLU 
Then, in the fine-tuning, \encoder{} helps retrieve topic-related data, which is later utilized for context enriching and merging with the help of trigger terms during the training of specific downstream tasks.
%with \encoder{}-retrieved 
%data and trigger terms 
%for various downstream-task fine-tuning.
%\encoder{} is pre-trained on 179 million hashtagged via contrastive learning to pull the same hashtags closer together in semantic space and push apart those with different hashtags.
Here we set up HCL with a \textit{\#Database} containing 45 million tweets grouped by hashtags for \encoder{} to retrieve context-enriching tweets.

To evaluate HICL's performance, we conducted experiments on seven popular Twitter benchmark datasets.
The main results demonstrate that HICL enables bidirectional language models, such as BART \cite{Lewis2019BARTDS}, RoBERTa \cite{roberta}, and BERTweet \cite{nguyen-etal-2020-bertweet}, to achieve superior performance by incorporating the top retrieved tweet from \encoder{}. 
Furthermore, inserting trigger terms between the source and retrieved tweets can enhance the overall performance, indicating that these trigger terms can positively facilitate information integration between the two components.

In further discussion of HICL, we first quantify the number of trigger terms and show that even a single trigger term can positively impact downstream tasks. 
Then, by probing into the position of trigger terms, we find those at the beginning or middle of sentences effectively facilitate information integration; in contrast, those at the end are less useful. 
Next, we quantify the scale of retrieved-context and observe augmenting more context is beneficial for enhancing social media NLU. However, the marginal benefits of adding additional text to the input diminish with the increasing number of retrieved pieces of information.
Finally, case studies and analysis of the trigger terms provide insight into how HICL helps NLU.

In summary, our contributions are three-fold:

%Our main contributions are summarized as follows:
$\bullet$ 
We propose a novel HICL framework for generic social media NLU in data sparsity, which can retrieve topic-related posts and enrich contexts via gradient-searched trigger terms.

%facilitates the incorporation of the retrieved text's context with gradient-searched trigger terms.

$\bullet$ We develop the first hashtag-driven pre-trained model, \encoder{}, leveraging hashtags to learn inter-post topic relevance (for retrieval) via contrastive learning over 179M tweets.

$\bullet$ We contribute a large corpus with 45M tweets for retrieval, and the experiments on 7 Twitter benchmarks show that HICL advances the overall results of various trendy NLU models.
% \footnote{We will make the HICL framework and benchmark with 45M tweets for retrieval and 7 public evaluation datasets open-source upon publication.}
\footnote{The HICL framework and benchmark with 45M tweets are available at\\ \url{https://github.com/albertan017/HICL}}

%We find HICL has promising empirical results on social media NLU and contribute a benchmark with 45M  (for retrieval) for future work in this direction.\footnote{The social media NLU benchmark will be open-source with \encoder{} and HICL upon publication.}

%% file: figure/intro-case.tex
\begin{figure}[t]
\centering
\includegraphics[width=0.45\textwidth]{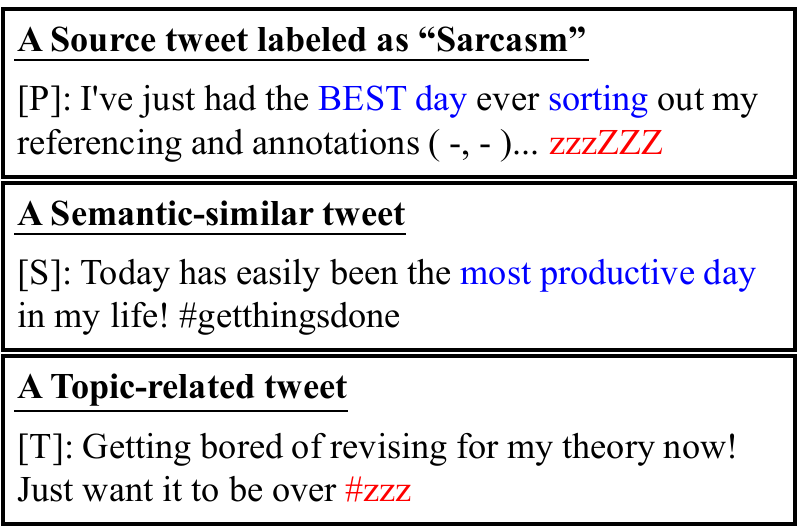}
 %\vspace{-0.5em}
\caption{\label{intro-case} 
A sample tweet $P$ sarcasing overwork through ``\textit{zzzZZZ}'' (top). 
It is followed by a SimCSE-retrieved tweet $S$ with similar semantics (middle) and a ``\textit{\#zzz}''-hashtagged tweet $T$ cross-refer other topic-related tweets (bottom). \textcolor{blue}{Blue words} show semantic indicators and \textcolor{red}{red words} topical hints. }
%\vspace{-1.5 em}
\end{figure}

%% file: 2Related-work.tex
\section{Related Work}

Our HICL is built upon in-context learning (ICL) and retrieves posts based on sentence embedding and hashtags.
In the following, we first discuss previous ICL work, followed by the discussion on sentence embedding and hashtag modeling. 

%involves extracting sentence embedding from a hashtag/topic perspective, enriching the context through in-context learning. Therefore our study is in line with sentence embedding, in-context learning, and hashtag, which we will discuss.

\paragraph{In-Context Learning} 

%In-context learning (ICL) was initially introduced in reference \cite{brown2020language}, where 
In the initial ICL work, researchers enhanced the GPT3 model's zero-shot inference potential by concatenating numerous exemplar instances ahead of the input text \cite{brown2020language}. 
It offers an interpretable interface for interacting with large language models (LLMs), making it easier to integrate human knowledge by modifying the templates and demonstrations. 
With the rapid scaling of LLMs size, the enormous computational expense of fine-tuning LLMs accentuates the necessity for ICL. 
To select good demonstration examples, researchers employed various metrics to retrieve samples, e.g., SentenceBert embeddings similarity \cite{liu-etal-2022-makes}, mutual information \cite{sorensen-etal-2022-information}, supervised retriever EPR \cite{rubin-etal-2022-learning}, etc.
There is also an inductive class learning experiment that showcases how demonstration samples drive end-task performance \cite{min2022rethinking}, which indicates that demonstration samples provide:
(1) instances from the label space demonstrating the range of possible labels,
(2) examples of the distribution of the input text, illustrating the kinds of inputs the model will encounter, and
(3) demonstrations of the overall format of the sequence, exhibiting the structure that the model's predictions should follow. 
These factors comprised the key reasons demonstration samples facilitated ICL model performance.

Although ICL has shown encouraging outcomes, previous work has predominantly concentrated on uni-directional models for natural language generation (NLG), such as GPT3 or LLaMa, leaving bi-directional models (such as the BERT family) largely unaddressed. 
Meanwhile, bi-directional models have shown unique advantages in NLU \cite{devlin-etal-2019-bert}.
It is because the bidirectional attention mechanisms can incorporate context from both directions when encoding a word or sentence, allowing effectiveness in capturing linguistic phenomena, such as long-distance dependencies, pronoun resolution, and negation understanding.
It also reflects how human readers process language since we understand words and sentences beyond solely relying on left-to-right contexts since it cannot fully capture the dependencies between the context words \cite{du-etal-2022-glm}.
%Nevertheless, ICL was initially designed to generate label words, making it uneasy to apply to fine-tune the bi-directional models.
%, which has proven effective in natural language understanding tasks.
We thus study tailor-making ICL to fine-tune bi-directional models and thoroughly evaluate its capabilities in social media NLU.

\paragraph{Sentence Embedding} 
Sentence embedding is the process of mapping sentences into continuous vector representations. 
It captures sentences' semantic meaning and allows them to be compared mathematically using distance metrics. 
This vector representation enables various downstream natural language processing (NLP) applications like sentence classification, semantic similarity, sentiment analysis, %\textcolor{blue}{
and is a widely-applied index in information retrieval.
%}.
Early work built sentence embeddings via averaging word vectors, e.g., word2vec \cite{word2vec}, which are word-level vector representations pre-trained from word co-occurrences.
Doc2Vec \cite{doc2vec} extended the idea of word embeddings to the document level and generated document embeddings by using either Distributed Memory mode or Distributed Bag of Words mode, where the former pre-trains embeddings by predicting words from their context and the latter do the opposite. Despite its simplicity, doc2vec has been shown to produce helpful sentence representations.

Inspired by siamese network, researchers later leveraged contrastive learning to obtain sentence embeddings. 
InferSent \cite{conneau-etal-2017-supervised} uses natural language inference (NLI) datasets to train a siamese bi-LSTM to predict the relations of input sentence pairs.
As the model is trained to distinguish between entailment, contradiction, and neutral relationships between sentence pairs, it forces the model to learn meaningful sentence representations. 
The idea of encoding sentences with the NLI dataset is further extended into transformer architecture in Universal Sentence Encoder \cite{cer2018universal}. 
And the corresponding results indicated that sentence embeddings are significantly helpful for transfer learning and can be used to obtain promising task performance with significantly less task-specific training data.
More recently, scholars have incorporated the concept of contrastive learning into the pre-training paradigm. 
SentenceBert \cite{reimers-gurevych-2019-sentence} is among the initial models to modify the pre-trained BERT model \cite{devlin-etal-2019-bert} by utilizing a siamese architecture to encode the semantic meaning of sentences into embeddings. 
SimCSE \cite{gao-etal-2021-simcse} presents an unsupervised method that utilizes standard dropout as noise and predicts an input sentence itself in a contrastive objective. 
They further include supervised contrastive learning with NLI datasets and reach state-of-the-art performance on semantic textual similarity (STS) tasks.
Although the dominant techniques for generating sentence embeddings are trained on formal written text such as the Standford Natural Language Inference dataset (SNLI) \cite{bowman-etal-2015-large} and Multi-Genre Natural Language Inference dataset (MNLI) \cite{williams-etal-2018-broad}, social media language - which is often characterized by sparsity and noise - has received relatively little attention. 
As a result, researchers have largely overlooked the informal writing style of social media language and instead adopted language encoders that are specifically designed for formal written text \cite{Wenzlitschke2022UsingBT,Tahaei2022IdentifyingAP}, which may compromise the final results.

As far as our understanding, there so far exist very few pre-trained models for sentence embedding that is specifically tailored for social media language. 
While some attempts have been made to pre-train language models on social media data, such as BERTweet \cite{nguyen-etal-2020-bertweet}, Bernice \cite{delucia-etal-2022-bernice}, and TwHIN-BERT \cite{zhang2022twhin}, most of them have been limited to using randomly-grouped tweets, which would result in a lack of coherent context and may consequently lead to confusion in pre-training. 
% \textcolor{blue}{Specially, TwHIN-BERT incorporates social engagements in Twitter into pre-training objective and embeds user behavior (e.g., like, reply, and retweet) into tweet representation. The TwHIN-BERT embeddings are helpful for detecting pairs of tweets appeal to similar users, but largely ignore topic information which is essential for social media language understanding under sparse condition.}
%\textcolor{blue}{
In contrast, our \#Encoder exhibits the first pre-trained sentence embedding model specifically tailored for social media language in a context-rich manner. 
Rather than prioritizing the semantic content of social media posts, usually characterized as noisy and lacking in context, \encoder{} adopts a topic-perspective view and utilizes hashtags as a means of grouping posts and driving contrastive pre-training for encoding social media posts.
Built upon the \encoder{}-learned embeddings, we further explore HICL, a novel framework on their use for downstream tasks under an in-context learning approach.

\paragraph{Hashtag Modeling} 
Our work is also related to prior studies using hashtags for language learning on social media platforms. Although social media language lacks context within individual posts, it offers a vast quantity of data. 
Hashtags, which are user-generated topic labels, are widely available on social media platforms and serve as clusters of post topics. 
These hashtags are typically used as indicators for constructing language resources \cite{van-hee-etal-2018-semeval,glandt-etal-2021-stance} and for social media tasks \cite{wang-etal-2019-topic,ding-etal-2020-hashtags,hashtag-topic-class}. 
For instance, hashtag semantics has been incorporated and benefit content recommendation \cite{weston-etal-2014-tagspace}.
Moreover, a recent study showed that adding automatically generated hashtags can enrich the context of tweets and help low resource classification \cite{Diao2023HashtagGuidedLT}.
% reference  proposed an approach to generate hashtags and add them to enrich the context of tweets. 
However, directly supplementing hashtags to tweets is arguably suboptimal as it may also bring noise and mislead the model because the appended hashtags and tweets may not be featured in the same semantic space for classification.
In contrast, to allow models to attend salient parts, we propose generating trigger terms to serve as a bridge for improving the integration between retrieved content and source input.
Moreover, they restricted their scope to low-resource classification with limited labeled data, whereas here, we focused on a more general scenario of social media NLU.
%their method relies on the hashtag generation model's results, restricting its application to general language models that have not been adequately pre-trained on social media corpora.
%Moreover, simple concatenation may be a suboptimal method to 
%Moreover, we notice directly supplementing retrieved information can distract the model.

In addition, some researchers work on hashtag embedding to help models gain hashtag-level understanding.
In this line,
%\textcolor{blue}{
%previous research has investigated methods to generate effective hashtag embeddings. 
Hashtag2Vec \cite{Liu2018Hashtag2VecLH} learns hashtag representations by jointly modeling their co-occurrence patterns and associated textual content; SHE \cite{Singh2020SHESH} captures semantic and sentiment information in hashtag embeddings leveraging multi-task learning.
%}
% More recently, reference \cite{Diao2023HashtagGuidedLT} proposed an approach to generate hashtags and add them to enrich the context of tweets. Their proposed method, HASHTATION, is a tweet classification model that employs hashtags as a guide. The model is intended to produce appropriate hashtags for a given tweet by collecting and encoding information from both post-level and entity-level data throughout the corpus. These generated hashtags can then be utilized as supplementary indicators for tweet classification. Despite demonstrating promising results, the effectiveness of the HASHTATION approach remains uncertain when applied to full datasets due to its setting on a low-source scenario (1\%-10\% training data). Moreover, the method heavily relies on the model's comprehension of hashtags, severely restricting its application to general language models that have not been adequately pre-trained on social media corpora.
Nevertheless, no prior work has exploited hashtags in gathering topic-related posts for large-scale language pre-training, which is a research gap we aim to address in this article. 
Here, we propose leveraging hashtags as topic indicators and employing contrastive learning to pre-train a \encoder{} model that can encode topic information. 
The \encoder{} model can then be utilized to retrieve topic-related posts and provide a concrete context that generic language models can interpret. 

%% file: 4Enrich.tex
\input{figure/retriever-train}

\section{HICL Framework}

This section introduces how we pre-train \encoder{} and apply it in the HICL framework. 
The framework design is first overviewed in Section \ref{ssec:framework-design}.
Then, we discuss the pre-training process for \encoder{} in Section \ref{ssec:pre-training} and how it is further leveraged in HICL to fine-tune language models in Section \ref{ssec:enrich}. Finally, we present the details to search for the trigger terms in Section \ref{ssec:trigger}. 
%The entire workflow is shown in Figure \ref{enrich-framework}.

%In this section, we present the HICL framework. We first brief the motivation for the HICL, then discuss the details on pre-training \encoder{}, building the \#Database, and retrieving the topic-related text. The overall HICL framework is presented in Figure \ref{enrich-framework}.

%\textcolor{red}{\subsection{Motivation}}
\subsection{Framework Design Overview\label{ssec:framework-design}} 

As discussed above, HICL employs \encoder{} for retrieving posts to enrich post-level context in task-specific fine-tuning.
For this reason, we feed \encoder{} with hashtag-grouped posts (posts with the same hashtag),
which differs from the BERTweet, Bernice or TwHIN-BERT scheme taking randomly concatenated tweets as input.
Our intuition is that posts about the same topic (hinted by hashtags) would allow richer context for pre-trained models to learn semantics. 
The grouping design considers that the limited words in a post may prevent the model's language learning potential from being fully exploited in pre-training.

%HICL framework originated from the observation that language patterns are distinct between pre-training corpus and fine-tuning datasets.

To better interpret this point, we first review the general design of most pre-trained models for NLU \cite{devlin-etal-2019-bert,roberta}.
It adopts a transformer encoder \cite{transformers} fed by a word sequence 
$\mathbf{x}=\langle x_1,x_2,...,x_L\rangle$ ($L$ is the word number).
For each word $x_i \in \mathbf{x}$ and its word embedding $e_i$, the model explores its representation $h_i$ through multiple self-attention encoder layers based on $x_i$'s occurrences with all words in $\mathbf{x}$.
A self-attention layer is formulated as follows:
%its representation $h_i$ in each encoder layer is 
%modeled with respect to all the other tokens in the sequence

\begin{equation}\small
h_i = \sum_{j=1}^{L}softmax(\frac{Q_iK_j}{d_k})V_i
\end{equation}

%Language modeling capability of pre-trained models can be attributed to the Masked Language Modeling \cite{devlin-etal-2019-bert,roberta} objective, 

% for each token $x_i$ and corresponding embedding $e_i$ in the input sequence, its representation $h_i$ in each encoder layer is modeled with respect to all the other tokens in the sequence from a self-attention perspective \cite{transformers}:
% \begin{equation}
% h_i = \sum_{j=1}^{L}softmax(\frac{Q_iK_j}{d_k})V_i
% \end{equation}
\noindent 
%where $L$ is the sequence length, 
$Q, K, V$ are projections of $\mathbf{x}$'s input embeddings. $d_k$ is the scaling factor to avoid a small gradient.

In pre-training, the transformer encoders tackle self-supervised learning tasks, e.g., masked language model (MLM), to explore the word features in context for learning general NLU skills.
However, because of the sparsely-distributed features, NLU encoders may need help to practice these tasks given post-level context only.
To mitigate sparsity, \encoder{} is pre-trained on grouped input with contrastive learning for a richer context in semantic learning.
Consequently, HICL matches a post with a retrieved post to follow this context-rich design and enable easier fine-tuning \cite{gururangan-etal-2020-dont}.

\subsection{\label{ssec:pre-training}\encoder{} Pre-training}

We then discuss how to pre-train \encoder{}, and the workflow is shown in Figure \ref{bert-train}.
It is built on the architecture of RoBERTa with a 12-layer transformer encoder \cite{transformers}.
%. We omit the architecture details and refer readers to \citet{transformers}.
We employ contrastive learning to pre-train large-scale tweets. 
In the following, we first discuss how to gather the pre-training data, followed by the training methods.

\paragraph{Pre-Training Data} 
\encoder{} is pre-trained on 15 GB of plain text from 179 million tweets and 4 billion tokens. Following the practice to pre-train BERTweet \cite{nguyen-etal-2020-bertweet}, the raw data was collected from the archived Twitter stream 
%grabbed by the Archive Team, \footnote{\url{https://archive.org/details/twitterstream}} 
containing 4TB of sampled tweets from January 2013 to June 2021.\footnote{\url{https://archive.org/details/twitterstream}} 
For data pre-processing, we ran the following steps.
First, we employed fastText \cite{joulin-etal-2017-bag} to extract English tweets and only kept tweets with hashtags.
Then, low-frequency hashtags appearing in less than 100 tweets were further filtered out to alleviate sparsity.
% After that, we obtained a large-scale dataset containing 179,065,725 unique tweets, each has at least one hashtag corresponding to 188,221 hashtags in total. 
After that, we obtained a large-scale dataset containing 179M tweets, each has at least one hashtag, and hence corresponds to 180K hashtags in total. 
% Finally, to set up pre-training, we randomly generated pairs of tweets sharing a hashtag for contrastive learning. 

%In the corpus, each hashtag contains on average 1091.3 tweets (a tweet may with multiple hashtags). 
\input{figure/bert-data-stat.tex}

To further examine how to utilize hashtags, we show the log-scaled distribution of hashtag frequency  
%Statistics for the number of tweets in each hashtag are presented 
in Figure \ref{bert-data-stat}. 
As can be seen, it is extremely imbalanced and roughly exhibits a long tail, where each hashtag appears in 951.4 tweets on average.
We observe that the majority (86\%) of hashtags contain less than 1,000 tweets, while several (the generic ones) appear in millions of tweets,
e.g., \#job occurs in 1.6 million tweets, \#nowplaying 1.3 million, and \#hiring 0.9 million. 

% For a more balanced training, tweet pairs from each hashtag are capped at 1,000 via sampling.
To enable a more balanced training, we sampled the posts with respect to the inverse of hashtag frequency and randomly formed pairs of tweets sharing a hashtag for contrastive learning. 
Besides, in order to guide \encoder{} to focus on non-trivial representation learning, we randomly add noise to hashtags, such as  deletion and segmentation \cite{hashformers}.
It is because hashtags are characterized by the \# symbol and the non-indent format, which may mislead the model to encode trivial and useless features for tackling pre-training tasks.

%we sample each hashtag with unique 1,000 pairs of tweets as positive instances to form the training corpus for contrastive pre-training. 
%Note that the model can easily recognize the hashtag, e.g. \#IloveMovie, the \# sign, and the non-indent format are typical indicators for hashtags. 
%To avoid the model learning trivial information during contrastive pre-training, i.e. only focusing on the hashtag, for the hashtags in the training corpus, we randomly delete, segment \cite{hashformers}, or keep the same. 

\paragraph{Pre-training Methods} 
To leverage hashtag-gathered context in pre-training, we exploit contrastive learning and train \encoder{} to identify pairwise posts sharing the same hashtag for gaining topic relevance. 
Formally,
%to encode the hashtag or topical information into sentence representation. 
given a batch of post pairs $D=\{[\mathbf{x}_1,\mathbf{x}_1^+],...,[\mathbf{x}_n,\mathbf{x}_n^+]\}$  ($\mathbf{x}_i$ and $\mathbf{x}_i^+$ are tagged the same hashtag), \encoder{} encodes $D$ into latent semantic space, $H=\{[h_1,h_1^+],...,[h_n,h_n^+]\}$ as their representations. Here the denotation was detailed in Section \ref{ssec:framework-design}.

In the hashtag-driven pre-training, \encoder{} aims to pull representations of posts with the same hashtag, $[h_i,h_i^+]$, closer and push apart those with different hashtags, $[h_i,h_j^+]$ ($i\neq j$).
%representations from different hashtags, e.g., $[h_1,h_2^+]$. 
Here we follow SimCSE \cite{gao-etal-2021-simcse} and compute the cross-entropy objective with in-batch negatives. And the training loss for a batch $D$ is defined as follows:
\begin{equation}\small\label{eq:contrastive-learning}
loss = -log\frac{e^{sim(h_i,h_i^+)/\tau}}{\sum_j^N{e^{sim(h_i,h_j^+)/\tau}}}
\end{equation}

\noindent where $sim(h_i,h_i^+)$ is the cosine similarity between post embedding $h_i$ and $h_i^+$, and $\tau$ is a temperature hyper-parameter.

To effectively encode the topic information, half of the input is constructed by concatenating posts with the same hashtag to the max sequence length, resulting in a single long document. 
The other half is present in individual posts. 
This way, \encoder{} can explore topic information in a context-rich setting while considering the limited length of social media posts.
Furthermore, we engage MLM with loss coefficient $\alpha$ as an auxiliary pre-training task to the aforementioned hashtag-driven contrastive learning.
It is to retain the word representation capability in \encoder{} pre-training.
%of the pre-training model. 

For evaluation of sentence encoding models on downstream tasks, we refer to SimCSE \cite{gao-etal-2021-simcse} and find that its results are inferior when directly fine-tuned for classification.
Likewise, \encoder{} is pre-trained on paired posts to learn topic relevance, which may better gain text-matching capability than classification.
We hence apply \encoder{} to retrieve posts in HICL fine-tuning, which will be discussed below.
% \textcolor{red}{
% Note that the \encoder{} is not for general-purpose social media NLU tasks. Similar to SimCSE, the \encoder{} target sentence representations but from a topic perspective. \citet{gao-etal-2021-simcse} notice that sentence-level objective may not directly benefit transfer tasks, i.e., fine-tuning on downstream tasks. We have the similar observation and do not include the \encoder{} for fine-tuning.
% }

% \input{figure/enrich-framework.tex}
\input{figure/HICL}
\subsection{\label{ssec:enrich}HICL Fine-tuning}

In fine-tuning, most NLU downstream tasks are formulated as a classification problem, which is to maximize posterior probability $P(y|\mathbf{x})$, meaning the most likely class $y$ given a post $\mathbf{x}$.
%formulated as predicting the probability of label $y$ given the input $x$:$P(y|x)$.
Due to data sparsity, the limited features in $\mathbf{x}$ may challenge NLU models to explicitly connect $\mathbf{x}$ to $y$.
%Unfortunately, classifying a social media post would be hard if the semantics connection between sample $x$ and label $y$ was indirect.
%Therefore, we 
We hence introduce a latent topic variable $z$ (from unlimited topic space on social media) to mitigate their information gap, and the theoretical formulation is as follows:
\begin{equation}\small
\begin{aligned}
% P(Y|X) &= \int_{z}P(Y|X,Z)P(Z|X)dZ \\
P(y|\mathbf{x}) &= \sum_{i}^{\infty}P(y\,|\,\mathbf{x},z_i)P(z_i\,|\,\mathbf{x})
\end{aligned}
\end{equation}

To estimate $P(y|\mathbf{x},z_i)$, it might be impractical to label extra data or enumerate all possible topics.
%However, it's impractical to enumerate all the possible topics, neither labeling extra data to estimate $P(y|x,z_i)$. 
%We propose to 
We then approximate the probability with the following steps:

    $\bullet$ Draw the most possible latent topic $z_i$ given the input $\mathbf{x}$ with formula $P(z_i|\mathbf{x)}$.
    
    $\bullet$ Retrieve a post from topic $z_i$ and concatenate it with  $\mathbf{x}$ to reflect the joint distribution of $P(\mathbf{x}, z_i)$. 
    
    $\bullet$ Model $P(y|\mathbf{x},z_i)$ via task-specific fine-tuning.
    %Modeling the classification probability $P(y|x,z_i)$ by language models.

% \noindent (1) Draw latent topic from prior $z_i ~ P(z_i|x)$.

% \noindent (2) Retrieve tweet sample from topic $z_i$ and concatenate it with the input $x$ to approximate the joint distribution of $P(x, z_i)$ 

% \noindent (3) Modeling the classification probability $P(y|x,z_i)$ by language models.

We design the following processes to run HICL fine-tuning and show the workflow in Figure \ref{enrich-framework}. 
First, \encoder{} was pre-trained to estimate the latent topic prior $P(z_i\,|\, \mathbf{x})$ with hashtag-driven contrastive learning (see Section \ref{ssec:pre-training}).
Then, for each post in the fine-tuning dataset, \encoder{} retrieves the most topic-related post based on $P(z_i\,|\, \mathbf{x})$. 
Next, we concatenate it with $\mathbf{x}$ to represent $P(\mathbf{x}, z_i)$.
In this way, the retrieved post may complement a view from a hashtag-indicated topic, enabling an enriched context for task-specific NLU learning.
%To that end, we propose the HICL framework, which (1) 
%leverages contrastive pre-training (\encoder{}) to estimate the latent topic prior $P(z_i|x)$; (2) search on the \#Database, retrieve the most similar post; and (3) concatenate with source input to approximate the joint distribution of $P(x, z_i)$ as well as simulate the long-sequence scenario in pre-training process. Language models consume the enriched text to conduct the final prediction $P(y|x,z_i)$.

%Therefore, context from topical perspective can provide an extra reference for the language model to digest and help understand the original post.

%\paragraph{Retrieval \#Database.} 
For the setup of a retrieval dataset, we consider the observations in Figure \ref{bert-data-stat}, where most hashtags have a hundred-scale frequency while very few million-scale.
%appear in hun, while a few hashtags are used by over one million tweets. 
%To maintain a reasonable search space, ensure class balance, and avoid overwhelming hashtags, 
To enable a reasonable search space for efficient and balanced retrieval, we randomly sampled at most 500 tweet samples from each hashtag group, resulting in 45 million unique tweets from 178,657 hashtags.
The dataset then bases the \encoder{}-retrieval in the HICL framework (thereby \#Database).

%\paragraph{Retrieval Method.}
For the retrieval method, we first encode a post $\mathbf{x}$ with \encoder{} to obtain its representation $h$. 
Here $\mathbf{x}$ can be any post with or without a hashtag.
Then, $h$ is matched with all posts' \encoder{}-encoded representation $h'$ 
in 
%and calculate its cosine similarity, $\frac{uv}{|u||v|}$, with 
\#Database to retrieve another post $\mathbf{x}'$, which results in the highest cosine similarity to $h$. $\mathbf{x}'$ is  considered as a topic-related post to $\mathbf{x}$ and concatenated it as enriched features for fine-tuning. 
%The entire HICL workflow is summarizes in Figure \ref{enrich-framework} for easy understanding.

%most topic-related tweet \textit{T}. The topic-related tweet \textit{T} and source tweet \textit{P} are 
%concatenated to form a context-enriched tweet for later use.

%\subsection{Enrich the Short Text}
% We first encode the source tweet \textit{P} with \encoder{} to obtain its topic embedding, and calculate its cosine similarity, $\frac{uv}{|u||v|}$, with respect to the \#Database to retrieve the most topic-related tweet \textit{T}. The topic-related tweet \textit{T} and source tweet \textit{P} are concatenated to form a context-enriched tweet for later use.

\subsection{Trigger Terms Search Algorithm\label{ssec:trigger}}

Here we further discuss how to fuse the retrieved and source context in fine-tuning. 
Although the retrieved posts are intended to provide supportive background, directly appending the two posts may be ineffective because the retrieved posts may not share the classification labels with the source and potentially confuse the model.
Accordingly, we propose inserting trigger terms optimized to combine the information from the retrieved text and source input, resulting in a coherent representation conducive to classification. 
Inspired by previous work \cite{zhong-etal-2021-factual}, we employ continuous vectors as trigger terms rather than utilizing natural language trigger terms. 
%which are human comprehensible and may facilitate interpretation, 
%we employ continuous vectors as triggers to draw inspiration from reference \cite{zhong-etal-2021-factual}. 
Concretely, given post $\mathbf{x}$, retrieved post $\mathbf{x'}$, and series of trigger terms $T_1, T_2,...,T_n$, we reformulate the input in the following form:
\begin{equation}
    [T_1,...,T_l], \mathbf{x}, [T_{(l+1)},...,T_m], \mathbf{x'}, [T_{(m+1)},...,T_n]
\end{equation}
For a reformulated input $\mathbf{x},\mathbf{x'},T$, the model's training loss is calculated as follows:
\begin{equation}
    \underset{\theta, T}{\textrm{argmin}} \ \mathcal L = -\sum \log P_{\theta}(y|\mathbf{x},\mathbf{x'},T)
\end{equation}
% \begin{equation}
%     \underset{\theta, T}{\textrm{argmax}} \ \mathcal L = \sum \log P_{\theta}(y|\mathbf{x},\mathbf{x'},T)
% \end{equation}
%\mathbf{}
To seek effective trigger terms, we first initialize trigger terms with random continuous embedding, and train the embeddings of the set of trigger terms, $T_1, T_2,...,T_n$, alongside other input tokens to establish a strong initialization. 
Following each iteration, we freeze the other model parameters and solely fine-tune the embeddings of these trigger terms for optimal solutions. 
%\textcolor{blue}{
We also present an ablation study on this iterative training process to evaluate its contributions (see Section \ref{sec:results}).
%}

% Concretely, given input $x$, retrieved text $x_r$, label $y$, the target is to search for trigger terms $x_t$, which can maximize the log-likelihood of training data:
% \begin{equation}
%     \underset{x_t}{\textrm{argmax}} \ \mathcal L(x,x_r,x_t) = \log p(y|e(x),e(x_r),e(x_{t}))
% \end{equation}
% where $e(x)$ represents the embedding of $x$. To search for the optimal trigger words $x_{t'}$, we add extra trigger tokens into the tokenizer, randomly initialize corresponding tokens embeddings and optimize them through the gradient.

%% file: figure/retriever-train.tex
\begin{figure*}[t]
\centering
\includegraphics[width=0.95\textwidth]{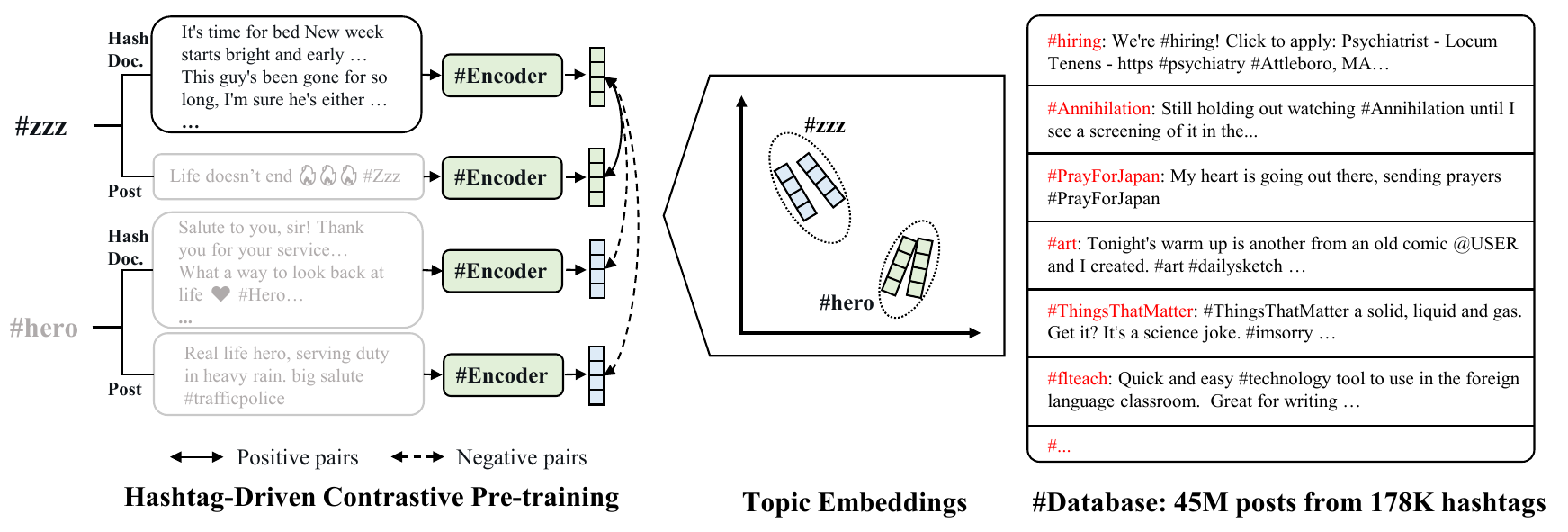}
% \vspace{-1em}
\caption{\label{bert-train}
The workflow to pre-train \encoder{} on 179M Twitter posts, each containing a hashtag. 
\#Retriever was pre-trained on pairwise posts, and contrastive learning guided them to learn topic relevance via learning to identify posts with the same hashtag.
We randomly noise the hashtags to avoid trivial representation.
%%and push apart those otherwise.
%\#BERT contrastive pre-training process. \#BERT takes the tweets from the same hashtag as positive instances and other in-batch samples as negative. \#BERT target at encoding the topic and presenting embeddings which can be distinguished with respect to hashtags.
}
% \vspace{-1em}
\end{figure*}

%% file: figure/bert-data-stat.tex
\begin{figure}[ht]
\centering
\includegraphics[width=0.45\textwidth]{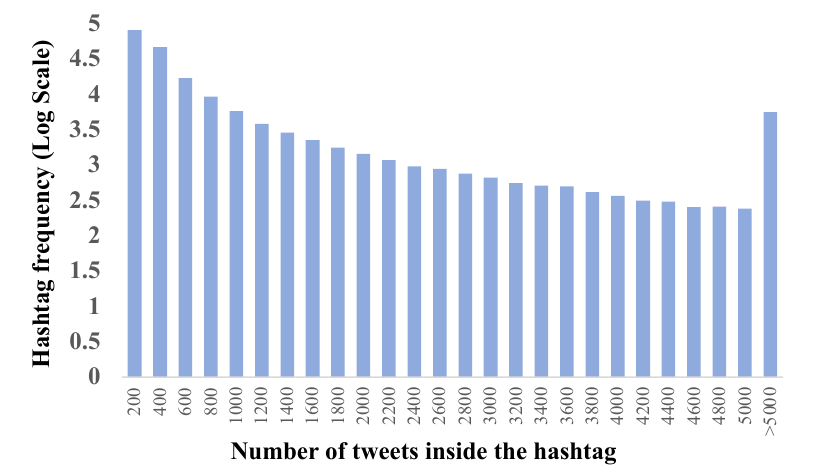}
% \vspace{-1em}
\caption{
Hashtag frequency distribution, which is imbalanced and exhibits a long tail. The x-axis shows the number of tweets presenting the hashtag and the y-axis the hashtag frequency in a log scale for better display.
%74\% appear in less than 500 tweets while 3\% in over 5,000 tweets.
%Statistics for tweets in hashtags in the pre-training corpus. The majority (74\%) of hashtags only contains less than 500 samples, while 3\% of hashtags are used by over 5,000 tweets. Such imbalance can hinder the training process and is resolved by only sampling 1000 unique pairs of tweets in each hashtag.
}
\label{bert-data-stat}
\vspace{-1em}
\end{figure}

%% file: figure/HICL.tex
\begin{figure*}[ht]
\centering
\includegraphics[width=0.9\textwidth]{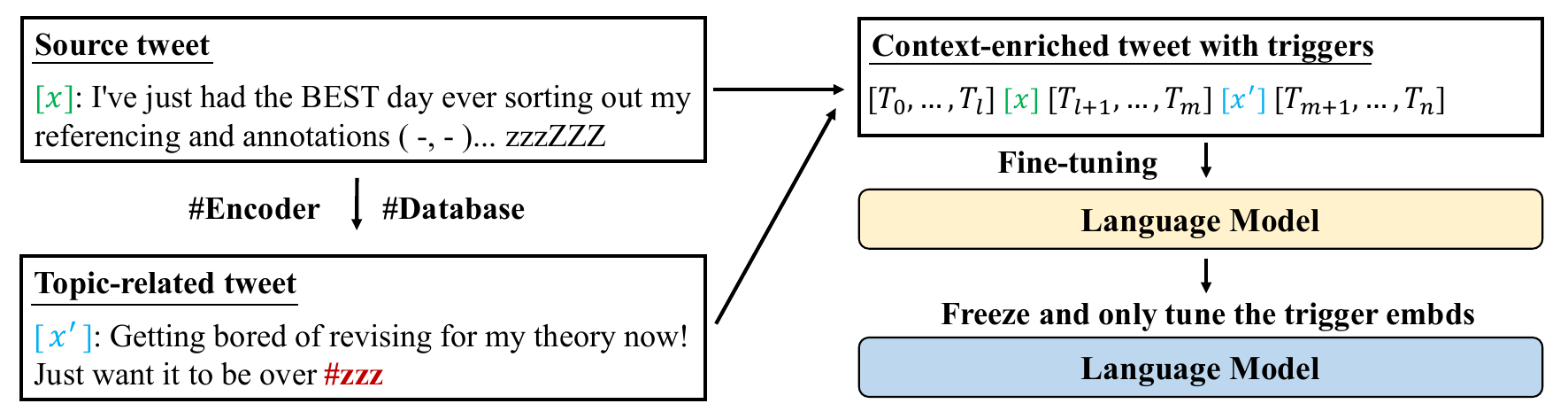}
% \vspace{-1em}
\caption{\label{enrich-framework}
The workflow of HICL fine-tuning.
A tweet $\mathbf{x}$ is first encoded with \encoder{} and the output is then used to search the \#Database to retrieve the most topic-related tweet $\mathbf{x}'$. 
After that, $\mathbf{x}'$ and $\mathbf{x}$ are paired in concatenation and inserted with trigger terms for task-specific fine-tuning. 
Here HICL can both work for tweets with and without hashtags.
%Note that source post does not required to have hashtags.
%The topic-related tweet and source tweet are then concatenated to generate a context-enriched tweet for later use.
}
% \vspace{-1em}
\end{figure*}

%% file: 5Experiment-setup.tex
\section{\label{sec:setup}Experimental Setup}

We set up the evaluation of HICL on the Twitter data, where we tested our fine-tuned results on 7 popular tasks to examine generic capability in social media NLU.
In the experimental discussion, a Twitter post is thereby referred to as a \textit{tweet}.

\paragraph{\encoder{} Pre-training Settings}
The hashtag-driven contrastive learning was implemented with Pytorch\footnote{\url{https://pytorch.org/}} and Hugging Face Transformers library\footnote{\url{https://github.com/huggingface/transformers}}. 
We primarily followed BERTweet configurations and initialized the \encoder{} parameters with BERTweet checkpoint for continued pre-training based on 4 NVIDIA RTX3090 GPUs. 
%We optimize the model using 
The pre-training was conducted by Adam optimizer with a peak learning rate set to 1e-5, maximum sequence length to 128, and batch size to 512.
We set temperate $\tau = 0.05$ in contrastive learning (as shown in Eq. \ref{eq:contrastive-learning}) and MLM loss coefficient $\alpha = 0.1$. 
\encoder{} was pre-trained for 10 epochs, roughly taking 7 days.

\paragraph{Benchmark Datasets}

% The evaluation was built upon 7 widely-used SemEval Twitter benchmarks, each concerning a different popular NLU task.
% They focus on the detection of
% \textit{\textbf{Emotion} } (SemEval-2018 Task1) \cite{mohammad-etal-2018-semeval}, 
% \textit{\textbf{Irony} }(SemEval-2018 Task3) \cite{van-hee-etal-2018-semeval}, 
% \textit{\textbf{Hate}}-speech (SemEval-2019 Task5) \cite{basile-etal-2019-semeval},
% \textit{\textbf{Offensive}} language (SemEval-2019 Task6) \cite{zampieri-etal-2019-semeval}, 
% and \textit{\textbf{Stance}}. 
% (SemEval-2016 Task6) \cite{mohammad-etal-2016-semeval}.
% These datasets were also included in TweetEval benchmark \cite{barbieri-etal-2020-tweeteval}.
% We also employed another two recent-year SemEval datasets: 
% \textit{\textbf{Humor}} detection (SemEval-2021 Task7) \cite{meaney-etal-2021-semeval}, 
% \textit{\textbf{Sarcasm}} detection (SemEval-2022 Task6) \cite{abu-farha-etal-2022-semeval}.

% Here \textit{Emotion} has four labels (Anger, Joy, Sadness, Optimism) and \textit{Stance} three (reflecting Favorable, Neutral, and Negative position towards a target). 
% The rest datasets adopt binary labels to indicate a ``yes-or-no'' in detection results.

The evaluation presented in this article is based on seven widely-used SemEval Twitter benchmark datasets, each related to a different popular natural language understanding (NLU) task. 
In the following, we briefly introduce each benchmark, and the corresponding statistics are presented in Table \ref{exp-setup-data-stat}.

\vspace{0.5em}
$\bullet$~\noindent\textbf{Stance Detection} focuses on understanding the author's stance and is formulated as follows. Given a tweet, the model aims to predict whether the author has a favorable, neutral, or unfavorable position toward a proposition or target. 
Here we employed the SemEval-2016 task 6 on Detecting Stance in Tweets, which provides five target domains: abortion, atheism, climate change, feminism, and Hillary Clinton. In this study, we merge the target domains and predict the stance.

\vspace{0.5em}
$\bullet$~\noindent\textbf{Emotion Recognition} is to recognize the author's emotion evoked by a tweet. We use SemEval-2018 task 1 dataset following TweetEval's practice, where the model should distinguish four emotions: anger, joy, sadness, and optimism.

\vspace{0.5em}
$\bullet$~\noindent\textbf{Irony Detection} focuses on recognizing whether a tweet includes ironic intents or not, making it a binary classification task. Here we used the data from SemEval-2018 task 3.

\vspace{0.5em}
$\bullet$~\noindent\textbf{Offensive Language Identification} aims to allow models to predict whether or not some offensive language is present in an input tweet, whose data is from SemEval-2019 task 6.

\vspace{0.5em}
$\bullet$~\noindent\textbf{Hate Speech Detection} is to predict whether a tweet is hateful against any of two target communities: immigrants and women. Our dataset comes from SemEval-2019 task 5.

\vspace{0.5em}
$\bullet$~\noindent\textbf{Humor detection} is to enable automatic detection of whether a given tweet exhibits a sense of humor, and the data is from the SemEval-2021 task 7.

\vspace{0.5em}
$\bullet$~\noindent\textbf{Sarcasm Detection} is a binary classification task of predicting whether a tweet shows a sense of sarcasm. The benchmark is set up based on SemEval-2022 task 6, which the tweet authors themselves labeled.

\vspace{0.5em}

Overall, these seven tweet classification benchmarks reflect a wide range of NLU capabilities to tackle social media data and comprehensively assess our proposed HICL framework's effectiveness in understanding such data.

\input{table/exp-setup-data-stat}

\paragraph{Comparison Setup}

We thoroughly experimented with the proposed HICL on the backbone of widely-employed bi-directional language models: BART and RoBERTa, and the state-of-the-art model for tweet NLU, BERTweet. In the following, we provide a concise overview of each model.

\vspace{0.5em}
\noindent$\bullet$~\textbf{BART} \cite{Lewis2019BARTDS} (Bidirectional and Auto-Regressive Transformer) is a pre-trained language model that employs the vanilla Transformer architecture. 
It can be viewed as a combination of the Bidirectional Encoder, similar to BERT, and an Autoregressive decoder, akin to GPT, into a single Seq2Seq model. 
BART is trained via a two-step process involving the corruption of text using an arbitrary noising function and the subsequent learning of a model to reconstruct the original text. 

\vspace{0.5em}
\noindent$\bullet$~\textbf{RoBERTa} \cite{roberta} (Robustly Optimized BERT Pretraining Approach) is an optimized BERT pre-training model through the use of larger data scales, longer training time, dynamic masking strategies, and optimized hyperparameters.

\vspace{0.5em}
\noindent$\bullet$~\textbf{BERTweet} \cite{nguyen-etal-2020-bertweet} is the first large-scale pre-trained model for the NLU of English tweets.
It leverages an 80GB corpus consisting of 850 million tweets. 
BERTweet adopts the RoBERTa architecture and training strategy yet concatenates tweets to achieve the maximum sequence length. Additionally, the model provides a specialized tokenizer for tweets.

\vspace{0.5em}

Our experiments consider taking these three models as the baselines to fine-tune the original datasets (namely \underline{Base}). 
For comparable results, HICL fine-tuning (see Section \ref{ssec:enrich}) was also carried out on varying base models, which takes paired input from a given tweet and its match retrieved by \encoder{}.
To allow the easy use of HICL, the pre-trained \encoder{} was directly applied for retrieval without task-specific fine-tuning.
Here we employed Faiss Library \cite{johnson2019billion} to speed up retrieval and costs around 30ms per 45M search on an Intel Xeon Gold 6248R CPU.
We empirically insert 5 trigger terms between the given tweet and its matched retrieved text.
Following this setup, we also examined HICL variants with pre-trained retrieval counterparts, enriching a tweet's context with SimCSE (namely \underline{SimCSE}).
We fine-tuned BERTweet for 30 epochs for each task with a warm-up learning rate of 1e-5 and batch size 16. 
We applied early stopping if no improvements were observed on validation for over 5 continuous epochs.
All models ran for 10 times, and we will report their average results in Section \ref{sec:results} below. 

In addition, we evaluate the effectiveness of conventional ICL, which involves conditioning the model's inferences on several demonstrations from training samples (namely \underline{ICL}). We follow the methods LMBFF proposed in \cite{gao-etal-2021-making} to implement this baseline. 
Concretely, we first sample a single example from each class for each input to create a demonstration set, and then perform prompt tuning to enable the model to learn from the demonstrations in the training set.

%% file: table/exp-setup-data-stat.tex
% \begin{table}[ht]
% \centering
% \includegraphics[width=0.45\textwidth]{figure/exp-setup-data-stat.pdf}
% % \vspace{-1em}
% \caption{Statistics for the data leakage in validation and testing sets with respect to the training set. We remove all the duplicates in the training set and further inspect it to build a clean training set.}
% \label{exp-setup-data-stat}
% \end{table}

\begin{table}[ht]
	\caption{
Benchmark dataset statistics.
	}
	\label{exp-setup-data-stat}
	 \centering
{\renewcommand{\arraystretch}{1.0}
\resizebox{0.3\textwidth}{!}
{
	\begin{tabular}[b]{l r rr }
		\toprule
		\textbf{Dataset}        & \textbf{Train} & \textbf{Val} & \textbf{Test} \\
		\midrule
		% \textbf{BERTweet} \\
		\textbf{Stance} 
		& 2,620 & 294 & 1,249 \\
        \textbf{Emotion} 
		& 3,257 & 374 & 1,421 \\
        \textbf{Irony} 
		& 2,862 & 955 & 784 \\
        \textbf{Offensive} 
		& 11,916 & 1,324 & 860 \\
        \textbf{Hate} 
		& 9,000 & 1,000 & 2,970 \\
        \textbf{Humor} 
		& 8,000 & 1,000 & 1,000 \\
        \textbf{Sarcasm} 
		& 3,114 & 353 & 1,400 \\
        
		% \hline
  %       \textbf{RoBERTa} \\
		% Baseline 
		% \hline
		% \textbf{Text+Image} \\
	 %    \textsc{ConcatFuse} & 52.86  & 81.62   & 34.78   & 39.19 &42.93 &71.09 \\
	 %    \textsc{Attention} & 54.30  & 82.64   & 33.71   & 39.23 & 39.41 & 71.48   \\
	 %    \textsc{Co-Attention} & 51.90  & 83.31   & 36.36   & 42.57 &40.59 &72.37  \\
	 %    \textsc{MultiheadAtt}& 53.69  & 84.33   & 36.96   & 42.11 &42.01 &73.33  \\

		\bottomrule	\end{tabular}}}
	% \vspace{-0.5em}

 % \vspace{-1em}
\end{table}

%% file: 6Experiment-results.tex
\input{table/exp-results-main}

\section{Experimental Results\label{sec:results}}

We first discuss the main comparison results and ablations in Section \ref{ssec:comparison}. 
Then, a quantitative analysis is presented in Section \ref{ssec:quan-analysis} to examine how trigger terms and retrieved tweets perform in varying scenarios, followed by a case study in Section \ref{ssec:case-study} to interpret how HICL benefits social media NLU. 
% Finally, we discuss more thoughts in Section \ref{ssec:discussions} to provide insight into future studies.

\subsection{Main Comparison Results and Ablation Study\label{ssec:comparison}}

The fine-tuned results on the 7 Twitter benchmarks (Section \ref{sec:setup}) and the averages are shown in Table \ref{exp-results-main}.

Our experimentation results provide support for our assertion that topic-related information, as obtained through the \encoder{} retrieved tweets, is more effective in enhancing generic NLU than semantic-related information (SimCSE-retrieved tweets) or demonstrations from similar training samples (+ICL). These results suggest that enriching a tweet's context with relevant tweets is a simple yet effective approach for improving generic NLU in data sparsity. As social media tweets face several sparsity problems, enriching the topic-related context becomes even more crucial in helping the language model understand the given scenario.
On the other hand, concatenating a semantic-similar tweet to the input may not be as helpful. 
%because it may not provide additional information or context to help the model make more accurate predictions. 
While a semantic-similar tweet may contain similar words or phrases to the input tweet, it may not necessarily provide additional context or information that can help the model better understand the topic being discussed. 

Besides, in-context learning is generally effective
%the basic ICL approach (LMBFF) shows a minor help
%the LMBFF approach (+ICL) 
%have been demonstrated to be effective 
in improving downstream task performance. 
This is done by providing demonstration tweets that are derived from training samples. 
These demonstrations could guide the model in NLU training and have been shown to improve the model's overall performance on downstream tasks. 
However, the degree of improvement achieved by the basic ICL or SimCSE is limited. The possible reason is that demonstration tweets from training samples are already familiar to the model or have been incorporated into its training. 
Meanwhile, HICL shows larger performance gains, implying that the topic-related tweets found by \encoder{} can better help the model comprehend the topic at hand and offers a relevant yet supplementary view. 
%and thus better collaborating with ICL in the HICL framework.
%Therefore, concatenating these tweets to the input tweet does not provide extra new information or context that can aid the model in better comprehending the topic at hand.

To further investigate the relative contributions of varying HICL modules, we present the ablation studies in Table \ref{exp-results-traintrigger}, where ``Base'' refers to the vanilla base models.
For other ablations,
we first examined the effectiveness of trigger terms, with ``+HICL w/o Tri.'' denoting simply concatenating the retrieved tweet with the source input.
Second, recall that in Section \ref{ssec:trigger}, we described that during training, we simultaneously train the embeddings of trigger terms and other tokens for initialization, followed by further fine-tuning the trigger embeddings after each iteration. 
An alternative approach would be to train the trigger embeddings jointly with the other token embeddings without additional tuning. 
We present comparative results to validate the importance of this additional tuning - ``+HICL w/o Add.'' indicates training without further tuning. ``+HICL'' denotes the full model with additional fine-tuning to optimize the trigger embeddings.

\input{table/exp-results-trainrigger}

The averaged results on 7 benchmarks are detailed in Table~\ref{exp-results-traintrigger}.
It demonstrates that inserting trigger terms between the source and retrieved tweets can enhance the final performance. 
Moreover, additional optimization of the trigger embeddings exhibits further downstream performance gains. 
These results support our hypothesis that trigger terms facilitate the merging of semantic information carried by the retrieved text and the source input. Thus, our study underscores the potential utility of trigger embeddings for generally improving the automatic NLU capability on social media language.
%of natural language processing systems.

\subsection{Quantitative Analysis\label{ssec:quan-analysis}}
In the previous section, we have shown the benefits of leveraging \encoder{}-retrieved tweets through our trigger term search algorithm.
In the following, we quantify how the trigger term usage and retrieved tweets help social media NLU learning.
The analyses for the sensitivity of trigger terms about their quantity and position will present first. 
Then, we examine the impact of the number of retrieved tweets on performance of the downstream tasks.

\input{table/exp-results-numtrigger}

\paragraph{Varying the Number of Trigger Terms}
Here we investigate how language models handle trigger terms with varying numbers and show the all-task average results in Table \ref{exp-results-numtrigger}, where ``T. \#$N$'' indicates $N$ trigger terms are inserted between the retrieved and source tweet.
We observe that although trigger terms are helpful, adding more trigger terms show minimal impact on the average performance. Notably, even a single trigger can positively affect the downstream task, reinforcing our argument that trigger terms are critical for facilitating the integration of information between the source and matched retrieved tweets.

\paragraph{Varying the Position of Trigger Terms}

In the previous experiments, the trigger term was empirically inserted between the source and retrieved tweets. 
We are then interested in 
%A further investigation was conducted into 
how the varying placement positions will result in the NLU learning outcome. 
The overall average results across all the tasks are illustrated in Table \ref{exp-results-placetrigger}. 
We utilize the terminology ``Front,'' ``Middle,'' ``End,'' and ``All'' to denote different trigger placements. 
Expressly, ``Front'' signifies the insertion of trigger terms before all the tweets, ``Middle'' refers to the placement of trigger terms between the source and retrieved tweets, and ``End'' represents the concatenation of trigger terms at the end. ``All'' denotes the inclusion of trigger terms in all of the positions above. ``No Trigger'' indicates that source and retrieved tweets are concatenated directly without triggers.

\input{table/exp-results-placetrigger}

Table \ref{exp-results-placetrigger} shows that trigger terms placed at the front or middle of tweets effectively facilitate information integration. 
In contrast, trigger terms placed at the end are generally unhelpful. 
It is intuitive why trigger terms in the middle produce the best results - their position provides explicit cues for connecting the source and retrieving information, acting as a ``bridge'' between the two. 
Trigger terms at the front also help, as they prime the language model to make a connection.
However, when placing the trigger terms at the end, the hint of such a ``connection'' may be weaker.
One possible explanation for this phenomenon is that placing trigger terms at the end may interfere with the natural sentence structure and disrupt the model's understanding of the input. 
The models are trained on data where relevant information is usually close to each other, which can bias the models to favor attending more strongly to adjacent or near-adjacent parts of the input. Therefore, placing the trigger terms at the end could cause the model to focus on resolving the unexpected input structure rather than integrating the source and retrieving information.

\paragraph{Varying Number of Retrieved Tweets}
We have analyzed the effects of trigger terms in fusing source and retrieved tweets. 
Then, we center on the retrieved tweets and examine how the number of retrieved tweets affects the performance. Figure \ref{exp-results-numpost} shows the all-task average results, where ``\#Enc.+$N$'' indicates top $N$ retrieved tweets are selected to concretize the context. We exclude the Sarcasm dataset for averaging due to its different trends and will discuss it later. 

\input{figure/exp-results-numpost}

The findings presented in Figure \ref{exp-results-numpost} suggest that augmenting the model's input with more contextual information generally enhances its NLU capabilities. 
However, for several reasons below, the marginal benefits of adding more text to the input gradually diminish with a continuous increase in the retrieved tweet number for use.
(1) \textit{Redundancy}: Concatenating multiple texts that revolve around the same topic may lead to redundancy in the input. This redundancy could limit the marginal utility of including the richer context in the input since the model may not obtain additional insights from repeatedly processing similar contents.
(2) \textit{Noise}: Adding more tweets to the input may introduce noise, as only part of the information may be task-relevant. This noise can hinder the model in identifying and concentrating on the most crucial information, thereby impeding performance gains.
(3) \textit{Model Capacity}: The capacity of a language model, which is determined by its architecture (e.g., number of layers, hidden units, and self-attention heads), may constrain its performance; 
even when more information is provided to the model by concatenating additional texts, the model may need the capacity to utilize this information to enhance its performance effectively.

\input{table/exp-results-numpostind}

To probe into the impact of the retrieved tweet number on individual tasks, we analyzed the slope of linear least squares while varying the number of retrieved tweets concerning task performance. The results are presented in Table \ref{exp-results-numpostind}.
Aside from the Sarcasm task, BART and RoBERTa typically exhibit performance gains as the number of concatenated tweets in the input increases for various tasks. In contrast, the BERTweet model does not enjoy such benefits due to its pre-training on randomly concatenated tweets, which lack coherence and hinder the model's ability to comprehend more extended context.
It is consistent with Figure \ref{exp-results-numpost}, where BERTweet presents flattened trends using more than 1 retrieved tweet, whereas BART and RoBERTa show a more apparent increasing trend.

%The suboptimal performance of the models on the 
Notably, the Sarcasm dataset negatively relates to a longer retrieved-context with all backbones. 
It can be attributed to the significant class imbalance, as only 24\% of the training data is labeled as sarcasm. This imbalance creates difficulties for the model in making accurate predictions, particularly under noisy conditions when concatenating more retrieved tweets.

\subsection{Qualitative Analysis \label{ssec:case-study}}

We have quantitatively shown how HICL benefits from using trigger terms and retrieved tweets.
Below, we qualitatively analyze some output of HICL to provide more insight into how it manages NLU learning on social media.

\paragraph{Trigger Terms}
We analyzed the Euclidean distance between the embeddings of trigger terms and all other tokens in the model vocabulary. 
Our findings indicate that trigger terms exhibit relatively smaller Euclidean distance and thus closer embedding similarity to the [mask], [pad], and [unk] tokens with respect to all other tokens in the vocabulary.

These special tokens, [mask], [pad], and [unk], have diffuse and indistinct semantic properties, as they function primarily as placeholders rather than conveyors of specific semantic content. 
Analogously, we posit that trigger terms improving model performance are likely to have similarly indistinct and diffuse semantic representations, as they act as placeholders or ``signal'' tokens, conveying information about the structural or intentional properties of the input rather than embedding precise semantic content. 
The semantic indeterminacy of these trigger terms may allow for a more flexible interpretation of the surrounding context and their use as placeholding signals %about input properties.
would further provide the model with useful structural information to improve downstream predictions. 
These results suggest why trigger terms are helpful in HICL design through a qualitative lens.

\paragraph{Retrieved Tweets}
%We have shown the superiority of HICL with \encoder{}. 
%To further study why HICL with \encoder{} exhibit superiority,  
For the usefulness of \encoder{}-retrieved tweets, we present some cases in Table \ref{exp-results-qualitative} to help interpret why it can benefit social media NLU. 
%\noindent\textbf{Case Study} is presented in 
For instance, the ``home alone'' in the first-row tweet is a movie's name, which may mislead the emotion detection model in predicting a negative emotion.
%the source tweet, \textit{@USER 3. home alone 4. fast and furious}, 
\encoder{} can connect it with other movie tweets through hashtag ``\#MovieTrivia'' to help NLU models cognize movie names to avoid errors in task-tackling.
Without such capability, SimCSE retrieved a tweet with similar words and offered limited help to make sense of movie names.

By qualitatively analyzing many cases, we find SimCSE tends to find tweets with similar words and sometimes cannot provide much extra information. 
On the contrary, \encoder{} can retrieve topic-related tweets, which may complement a %slightly 
different view to gain topic-level knowledge for better NLU.
\input{table/exp-results-case}

%is aware of the movie names, ``Home Alone'' and ``Fast and Furious'', retrieves a tweet referring to the topic \#Movie, concretes the source text and helps the final prediction.

% While the semantic similar tweet, \textit{home alone.. \#FreakingOut} further inflates the sad emotion. On the contrary, a topic-related tweet from \encoder{} is aware of the movie names, ``Home Alone'' and ``Fast and Furious'', retrieves a tweet referring to the topic \#Movie, concretes the source text and helps the final prediction.

%% file: table/exp-results-main.tex
\begin{table*}[t]

\caption{
Comparison results of different models with varying bi-directional backbones. The best results in each column under a backbone are \underline{underlined}. Our HICL framework significantly outperforms other comparison models on average with $p<0.05$.
}
	 \centering
{\renewcommand{\arraystretch}{1.0}
\resizebox{1.0\textwidth}{!}
{
	\begin{tabular}[b]{l r r r r r r r r}
		\toprule
		\textbf{Method}        & \textbf{Stance} & \textbf{Emotion} &\textbf{Irony} & \textbf{Offensive} & \textbf{Hate} 
        & \textbf{Humor}  & \textbf{Sarcasm} & \textbf{Average}
        \\
		\midrule
		\textbf{BART} \\
		Base 
		& $67.3\pm0.6$ & $77.8\pm0.5$ & $67.3\pm1.9$ & \underline{$81.2\pm0.6$} & $49.4\pm1.8$ & $95.2\pm0.3$ & \underline{$34.6\pm1.6$} & $67.5\pm1.1$ \\
        +ICL
        & $67.4\pm1.8$ & $77.4\pm0.9$ & \underline{$69.5\pm1.1$} & $80.8\pm0.8$ & $50.2\pm1.4$ & $94.4\pm0.4$ & $33.3\pm1.7$ & $67.6\pm1.1$ \\
        +SimCSE
        & $66.3\pm1.4$ & $76.3\pm0.4$ & $66.4\pm2.4$ & $79.6\pm1.1$ & $51.0\pm1.8$ & \underline{$95.3\pm0.3$} & $32.7\pm1.2$ & $66.8\pm1.2$ \\
        % HICL w/o Trigger
        % & \underline{$68.1\pm1.1$} & $77.6\pm0.4$ & $68.2\pm1.5$ & \underline{$81.2\pm0.5$} & \underline{$51.4\pm2.4$} & $94.4\pm0.5$ & $34.0\pm1.7$ & $67.8\pm1.1$ \\
        +HICL
        & \underline{$68.0\pm1.0$} & \underline{$78.6\pm0.4$} & $68.6\pm0.8$ & $80.9\pm0.9$ & \underline{$51.0\pm1.1$} & $94.7\pm0.4$ & $34.5\pm2.5$ & \underline{$68.1\pm1.0$} \\
        % +\#BERT + Trigger
        % & \underline{$69.4\pm0.6$} & $80.9\pm0.5$ & $79.5\pm0.7$ & $79.3\pm1.0$ & \underline{$58.8\pm2.3$}
        % & $95.9\pm0.0$ & \underline{$44.8\pm2.5$} & \underline{$72.7\pm1.0$} \\
        \midrule
		% \hline
        \textbf{RoBERTa} \\
        Base 
		& $69.0\pm0.5$ & $78.2\pm0.5$ & $64.3\pm2.6$ & $79.7\pm0.9$ & $47.9\pm1.8$ & \underline{$95.0\pm0.6$} & $38.0\pm2.5$ & $67.4\pm1.4$ \\
        +ICL
        & $67.5\pm1.4$ & $77.8\pm0.7$ & $68.6\pm1.8$ & $79.5\pm1.2$ & $50.8\pm1.3$ & $94.2\pm0.4$ & $36.0\pm1.9$ & $67.8\pm1.2$ \\
        +SimCSE
        & $68.0\pm0.7$ & $77.1\pm1.0$ & $68.8\pm2.3$ & $78.5\pm1.0$ & $48.6\pm1.8$ & $94.9\pm0.3$ & $36.8\pm1.8$ & $67.5\pm1.3$ \\
        % HICL w/o Trigger
        % & \underline{$69.3\pm1.0$} & $77.7\pm0.8$ & \underline{$72.9\pm1.0$} & $78.4\pm1.0$ & \underline{$51.3\pm2.2$} & $94.8\pm0.4$ & $38.0\pm1.1$ & $68.9\pm1.1$ \\
        +HICL
        & \underline{$69.4\pm1.3$} & \underline{$78.4\pm0.6$} & \underline{$72.8\pm1.8$} & \underline{$79.9\pm0.7$} & \underline{$51.2\pm1.4$} & $94.7\pm0.2$ & \underline{$41.0\pm2.1$} & \underline{$69.6\pm1.2$} \\
        \midrule
		% \hline
        \textbf{BERTweet} \\
        Base 
		& \underline{$70.3\pm0.9$} & $81.2\pm0.8$ & $78.7\pm1.4$ & \underline{$80.5\pm0.8$} & $54.9\pm0.9$ & $95.9\pm0.3$ & $45.9\pm2.7$ & $72.5\pm1.1$ \\
        +ICL
        & $69.8\pm1.5$ & $67.5\pm0.9$ & $80.3\pm1.4$ & $76.4\pm1.3$ & \underline{$58.6\pm2.2$} & $94.4\pm0.6$ & $43.3\pm0.8$ & $70.0\pm1.3$ \\
        +SimCSE
        & $69.0\pm0.8$ & $80.5\pm0.7$ & $80.9\pm1.5$ & $80.1\pm0.7$ & $56.5\pm1.6$ & \underline{$96.2\pm0.4$} & $47.2\pm1.9$ & $72.9\pm1.1$ \\
        % HICL w/o Trigger
        % & $69.0\pm1.0$ & \underline{$81.6\pm0.4$} & $79.9\pm1.4$ & $79.8\pm0.3$ & \underline{$56.7\pm2.2$} & $96.0\pm0.4$ & $47.8\pm0.6$ & $73.0\pm0.9$ \\
        +HICL
        & $69.5\pm0.7$ & \underline{$81.2\pm0.6$} & \underline{$81.5\pm0.9$} & $80.1\pm0.6$ & $56.1\pm1.8$ & $96.0\pm0.3$ & \underline{$49.0\pm2.4$} & \underline{$73.4\pm1.1$} \\

		\bottomrule	
        \end{tabular}
        }
        }
	
	% \vspace{-1em}
	\label{exp-results-main}
\end{table*}

%% file: table/exp-results-trainrigger.tex
\begin{table}[t]
\caption{
    Average results for different training methods.
}
% \vspace{-1em}
\label{exp-results-traintrigger}
	 \centering
{\renewcommand{\arraystretch}{1.0}
\resizebox{0.43\textwidth}{!}
{
	\begin{tabular}[b]{l r r r r}
		\toprule
		\textbf{model}  & \textbf{Base} & \textbf{+HICL w/o Tri.} & \textbf{+HICL w/o Add.} &\textbf{+HICL} \\
		\midrule
		% \textbf{BERTweet} \\
		\textbf{BART} 
		& 67.5 & 67.8 & 67.5 & 68.1  \\
        \textbf{RoBERTa} 
		& 67.4 & 68.9 & 68.8 & 69.6  \\
        \textbf{BERTweet} 
		& 72.5 & 73.0 & 73.4 & 73.4  \\
        \textbf{Average} 
		& 69.2 & 69.9 & 70.0 & 70.3  \\
        
		\bottomrule	\end{tabular}}}
	\vspace{-0.5em}

 % \vspace{-1em}
\end{table}

%% file: table/exp-results-numtrigger.tex
\begin{table}[t]
\caption{
    Average results varying the number of triggers.
}
% \vspace{-1em}
\label{exp-results-numtrigger}
	 \centering
{\renewcommand{\arraystretch}{1.0}
\resizebox{0.48\textwidth}{!}
{
	\begin{tabular}[b]{l r r r r r r r}
		\toprule
		\textbf{model} & \textbf{Base} & \textbf{+HICL w/o Tri.} & \textbf{T. \#1} & \textbf{T. \#3} &\textbf{T. \#5} & \textbf{T. \#7} &\textbf{T. \#9} \\
		\midrule
		% \textbf{BERTweet} \\
		\textbf{BART} 
		& 67.5 & 67.8 & 67.9 & 68.0 & 68.0 & 68.2 & 68.0  \\
        \textbf{RoBERTa} 
		& 67.4 & 68.9 & 69.0 & 68.9 & 69.6 & 68.7 & 68.9  \\
        \textbf{BERTweet} 
		& 72.5 & 73.0 & 73.2 & 73.5 & 73.4 & 73.3 & 73.4 \\
        \textbf{Average} 
		& 69.2 & 69.9 & 70.0 & 70.1 & 70.3 & 70.1 & 70.1 \\
        
		\bottomrule	\end{tabular}}}
	% \vspace{-0.5em}

 % \vspace{-1em}
\end{table}

%% file: table/exp-results-placetrigger.tex
\begin{table}[t]
\caption{
    Average results varying the placing positions of triggers.
}
% \vspace{-1em}
\label{exp-results-placetrigger}
	 \centering
{\renewcommand{\arraystretch}{1.2}
\resizebox{0.48\textwidth}{!}
{
	\begin{tabular}[b]{l r r r r r r}
		\toprule
		\textbf{model}  & \textbf{Base} & \textbf{No Trigger} &\textbf{Front} & \textbf{Middle} &\textbf{End} &\textbf{ALL}  \\
		\midrule
		% \textbf{BERTweet} \\
		\textbf{BART} 
		& 67.5 & 67.8 & 67.7 & 68.1 & 67.7 & 68.0 \\
        \textbf{RoBERTa} 
		& 67.4 & 68.9 & 69.1 & 69.6 & 68.6 & 69.0\\
        \textbf{BERTweet} 
		& 72.5 & 73.0 & 73.7 & 73.4 & 73.0 & 73.4\\
        \textbf{Average} 
		& 69.2 & 69.9 & 70.2 & 70.3 & 69.8 & 70.1\\
        
		\bottomrule	\end{tabular}}}
	% \vspace{-0.5em}

 % \vspace{-1em}
\end{table}

%% file: figure/exp-results-numpost.tex
\begin{figure}[t]
\centering
\includegraphics[width=0.48\textwidth]{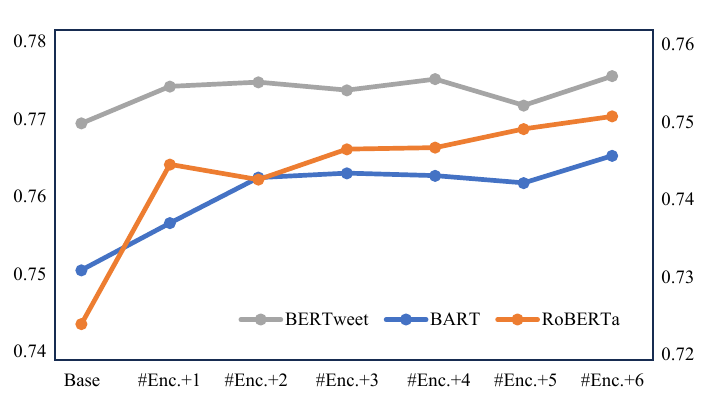}
% \vspace{-1em}
\caption{
Average results on varying the number of retrieved tweets, excluding the Sarcasm dataset because of its different trends (to be discussed separately).
}
\label{exp-results-numpost}
% \vspace{-1em}
\end{figure}

%% file: table/exp-results-numpostind.tex
\begin{table}[t]
\caption{
    Linear Least Squares Regression Slope with Number of Retrieved Tweets as the Independent Variable and Task Performance as the Dependent Variable, Coefficients Multiplied by 1000 for Presentation Purposes.
}
% \vspace{-1em}
\label{exp-results-numpostind}
	 \centering
\setlength{\tabcolsep}{4pt}
{\renewcommand{\arraystretch}{1.2}
\resizebox{0.48\textwidth}{!}
{
	\begin{tabular}[b]{l r r r r r r r r}
		\toprule
		% \textbf{Method} & \textbf{Stance} & \textbf{Emotion} &\textbf{Irony} & \textbf{Offensive} & \textbf{Hate} 
  %       & \textbf{Humor}  & \textbf{Sarcasm} & \textbf{Average} \\
        \textbf{Method} & \textbf{Sta.} & \textbf{Emo.} &\textbf{Iroy} & \textbf{Off.} & \textbf{Hate} 
        & \textbf{Hum.}  & \textbf{Sar.} & \textbf{Avg} \\
		\midrule
		% \textbf{BERTweet} \\
		\textbf{BART} 
		& 0.67 & 2.66 & 3.49 & 0.32 & 0.08 & 0.01 & -6.79 & 0.04  \\
        \textbf{RoBERTa} 
		& 1.66 & 2.72 & 1.63 & -1.47 & 3.26 & 0.92 & -7.37 & 0.19  \\
        \textbf{BERTweet} 
		& 1.35 & -0.02 & -1.65 & 0.25 & -0.03 & -0.05 & -2.02 & -0.31  \\
        
		\bottomrule	\end{tabular}}}
	% \vspace{-0.5em}

 % \vspace{-1em}
\end{table}

%% file: table/exp-results-case.tex
% \begin{figure}[ht]
% \centering
% \includegraphics[width=0.5\textwidth]{figure/exp-results-qualitative.pdf}
% % \vspace{-1em}
% \caption{exp-results-qualitative.}
% \label{exp-results-qualitative}
% \end{figure}

\begin{table*}[tb]
	\caption{
 Three cases from Emotion, Stance, and Sarcasm datasets. The columns from left to right show task, source tweet (for retrieval), semantic-similar tweet (retrieved by SimCSE), and topic-related tweet (retrieved by \encoder{}).
	}
	 \centering
\small%, \normalsize, \large, \Large
{\renewcommand{\arraystretch}{1.3}
\resizebox{1.0\textwidth}{!}
{
	\begin{tabular}[b]{p{0.07\textwidth} p{0.3\textwidth} p{0.3\textwidth} p{0.3\textwidth}}
		\toprule
		\textbf{Task} & \textbf{Source tweet}        & \textbf{Semantic-similar tweet} & \textbf{Topic-related tweet} 
        \\
		\midrule
		% \textbf{BERTweet} \\
		Emotion: Joy &
        @USER 3. home alone 4. fast and furious &
        home alone.. \#FreakingOut &
        @USER The Amazing Spiderman \#MovieTrivia \\

        \hline
        Stance: \newline Favor &
        \#Mission : \#Climate @USER home $>$ Run your dishwasher only if it's full. ( by @USER \#Tip \#ActOnClimate \#SemST &
        Only do laundry when you have a full load. The same holds true with your dishwasher. Only run when full. \#moneysavingtips \#energysavers &
        Learn how we're making homes and \#buildings more energy efficient than ever. HTTPURL \#ActOnClimate HTTPURL\\

        \hline
        Sarcasm &
        I love a Monday morning so glad the weekends over! &
        Loves a Monday morning \#whaaaaat &
        How is it half 8 already?? \#hatemondays \#weekendplease \\

		% \hline
  %       \textbf{RoBERTa} \\
		% Baseline 
		% \hline
		% \textbf{Text+Image} \\
	 %    \textsc{ConcatFuse} & 52.86  & 81.62   & 34.78   & 39.19 &42.93 &71.09 \\
	 %    \textsc{Attention} & 54.30  & 82.64   & 33.71   & 39.23 & 39.41 & 71.48   \\
	 %    \textsc{Co-Attention} & 51.90  & 83.31   & 36.36   & 42.57 &40.59 &72.37  \\
	 %    \textsc{MultiheadAtt}& 53.69  & 84.33   & 36.96   & 42.11 &42.01 &73.33  \\

		\bottomrule	\end{tabular}}}
	% \vspace{-1em}

	% \vspace{-1em}
	\label{exp-results-qualitative}
\end{table*}

%% file: 7Conclusions.tex
\section{Conclusions}
We have proposed a hashtag-driven in-context learning (HICL) framework with a pre-trained \encoder{} based on hashtags to retrieve topic-related social media posts, which are combined with the source input for context enriching via gradient-optimized trigger terms for task-specific fine-tuning. 
\encoder{} is pre-trained on 179 million hashtagged tweets using contrastive learning, enabling it to associate tweets with matching hashtags and differentiate those with divergent topics. 
We implemented HICL with a \textit{\#Database} of 45 million hashtag-grouped tweets, allowing \encoder{} to acquire and integrate context with triggers in task-specific fine-tuning.

We conducted experiments on 7 widely-used Twitter benchmark datasets to evaluate \encoder{} and HICL's effectiveness. Our results indicate that HICL significantly enhances the performance of bidirectional language models such as BART, RoBERTa, and BERTweet by incorporating the top-retrieved tweets from \encoder{}. 
Additionally, we found that incorporating trigger terms between the source and retrieved tweets can improve overall performance, suggesting that trigger terms facilitate effective information integration.

Through a quantitative analysis of trigger terms, we demonstrated that even a single trigger can positively influence downstream tasks. Further investigation revealed that trigger terms at the beginning or middle of sentences contribute to effective information integration, whereas those positioned at the end of sentences are generally less beneficial. Moreover, supplementing the model with additional context improves language comprehension abilities, although the marginal benefits decrease as more information is retrieved.

Despite the promising results of the HICL framework, it presents several limitations requiring future research:

Firstly, our pre-training corpus relies on abundant user-annotated hashtags, which lack quality assurance. Additionally, hashtag frequency exhibits a long-tail distribution, leading to class imbalance challenges. Investigating automatic methods to create a high-quality pre-training corpus could be valuable.

Secondly, our retrieval method utilizes a large \#Database with 45 million tweets and requires 30ms for retrieval on an Intel Xeon Gold 6248R CPU. Corpus distillation techniques, such as clustering and indexing, could improve retrieval efficiency while maintaining acceptable performance levels.

Thirdly, the HICL framework and \encoder{} do not enforce semantic consistency during retrieval. 
Although our experiments have validated the effectiveness of the proposed framework, extra efforts in selecting the optimal context through re-ranking algorithms can allow more performance gain and provide a better solution to the data-sparsity challenge.